\documentclass[conference]{IEEEtran}
\IEEEoverridecommandlockouts

\usepackage{amsmath,amssymb,amsfonts}
\usepackage{algorithmic}
\usepackage{graphicx}
\usepackage{textcomp}
\usepackage{xcolor}
\usepackage{authblk}
\usepackage{cite}
\usepackage{url}
\usepackage[hidelinks,colorlinks=true,linkcolor=blue,citecolor=green,urlcolor=blue]{hyperref}
\def\BibTeX{{\rm B\kern-.05em{\sc i\kern-.025em b}\kern-.08em
    T\kern-.1667em\lower.7ex\hbox{E}\kern-.125emX}}
    
\begin{document}

\title{Ratatouille: A tool for Novel Recipe Generation}

\author[1]{Mansi Goel\textsuperscript{$\dagger$}}
\author[2]{Pallab Chakraborty\textsuperscript{$\dagger$}}
\author[2]{Vijay Ponnaganti\textsuperscript{$\dagger$}}
\author[2]{Minnet Khan}
\author[2]{\\Sritanaya Tatipamala}
\author[2]{Aakanksha Saini}
\author[*]{Ganesh Bagler\textsuperscript{1}}
\affil[1]{\textit{\ Center for Computational Biology, Indraprastha Institute of Information Technology (IIIT-Delhi), New Delhi, India}}
\affil[2]{\textit{\ Department of Computer Science, Indraprastha Institute of Information Technology (IIIT-Delhi), New Delhi, India}}
\affil[*]{\textit{\ Corresponding author: Ganesh Bagler, bagler@iiitd.ac.in}}

\renewcommand\Authands{ and }

\maketitle

\def\thefootnote{$\dagger$}\footnotetext{These authors contributed equally to this work}

\begin{abstract}
Due to availability of a large amount of cooking recipes online, there is a growing interest in using this as data to create novel recipes. Novel Recipe Generation is a problem in the field of Natural Language Processing in which our main interest is to generate realistic, novel cooking recipes. To come up with such novel recipes, we trained various Deep Learning models such as LSTMs and GPT-2 with a large amount of recipe data. We present Ratatouille (\url{https://cosylab.iiitd.edu.in/ratatouille2/}), a web based application to generate novel recipes. 
\end{abstract}

\section{Introduction}
Novel recipe generation is the task of generating new recipes from given ingredients. It is a natural language processing task and falls under semi-structured text generation. Often cooking recipes are divided into sections like ingredients, utensils/appliances required, along with a set of instructions, which is why a recipe's content can be considered as a structured piece of text. The problem of automatic generation of cooking recipes is exciting and challenging since recipes need to generate ingredients and cooking instructions based on the preferred list of ingredients given by the user. Text generation uses computational linguistics and machine learning to create such recipes. The evaluation of such recipes is still not trivial, and existing metrics only provide a partial picture when it comes to evaluating the quality of the generated recipes.\\\par

Deep learning models are trained with the anticipation of novel recipe generation through the list of ingredients. Several models were propounded Transformer-based model \cite{h2020recipegpt}, routing enforced generation model \cite{yu2020routing}, language model \cite{parvez2018building,kiddon2016globally,agarwal2020building}, knowledge-based models \cite{koncel2019text,chen2019distilling} and GRU (Gated Recurrent Unit) based encoder-decoder model \cite{majumder-etal-2019,bosselut2018discourse}. These models were able to generate novel recipes along with recipe instructions and ingredients list but most of them could not generate quantity and units of ingredient. Some of the models also lacked context and dismissed the inputs from the user.  
\par

This article deals with the generation of novel recipes given the list of ingredients using deep learning models (character-level LSTM, word-level LSTM model and transformer-based pre-trained GPT-2 model). The models are created through Open-AI, trained on diverse topics from 8 million web pages and contain 1.5 billion parameters to make text predictions. We train our model using RecipeDB dataset~\cite{batra2020recipedb} to generate a corpus of recipes containing title, ingredients, and cooking instructions. Our objective is that, based on the input list of ingredients, model generates novel and diverse recipes. \\

Recipe generation is quite challenging problem, recipe's quality is of great importance. We used BLEU score to determine the quality of recipes. This work is also available as a web application, Ratatouille (\url{https://cosylab.iiitd.edu.in/ratatouille2/}).\\

This article is divided into multiple sections. The brief review of related work is available in Section II. Section III explains about the dataset and our methodology for novel recipe generation. Section IV demonstrate about our web application. In the end, the conclusion and future work is explained in Section V. 

\section{Related Work}
Several advances in the field of computation and dataset availability allows to generate recipes from food images \cite{salvador2019inverse, wang2020structure,liang2015recurrent} and generate personalized recipes based on user demands~\cite{majumder-etal-2019, yu2020routing}. A rule-based generation algorithm has been implemented to develop a recipe generation system, EPICURE~\cite{dale1990generating}. Salvador et al.~\cite{salvador2019inverse} used newly released Recipe1M+ ~\cite{marin2019recipe1m+} large-scale dataset to generate the recipes from food images and evaluate the recipes using perplexity score. \\

Cooking recipes are semi-structured text data containing recipe title, ingredients, and cooking instructions. Due to the recent developments in the field of natural language generation, multiple state-of-the-art models have been released, but those models are still inefficient in generating context-based and structured text. Language models trained on large-scale data generate good human-readable text, but these models cannot be constrained to generate any specific aspect of some document. CTRL~\cite{keskar2019ctrl}, a conditional transformer model consisting of 1.63 billion parameters, was trained with conditional tokens which helps in controlling the generation for any specific domain, entities, dates, subdomains. This model can also be used for generating context-specific keywords from a document which it uses internally for selecting the most appropriate word for generating the desired sequence.\\

In transfer learning approach the model is first trained on a large dataset and then used for multiple other downstream tasks by fine-tuning the pre-trained model. Every text processing problem can be categorized as a text-to-text task~\cite{raffel2019exploring} where text is fed as input to the model, which then generates the target text. Several efforts have been made to utilize language modelling over recipe datasets. Parvez et al.~\cite{parvez2018building} used `Now You're Cooking' dataset containing 150,000 recipes to construct a LSTM-based language model. Based on the research of~\cite{parvez2018building}, Agarwal et al. ~\cite{agarwal2020building} created a hierarchically disentangled language model by curating 158,473 recipes using named-entity recognition and unsupervised methods. Kiddon et. al.~\cite{kiddon2016globally} proposed an encoder-decoder based neural checklist model using 84,000 recipes to generate recipes by maintaining a checklist of accessible and already used ingredients. With the similar spirit, Lee et al. ~\cite{h2020recipegpt} presented a transformer-based GPT-2 model (RecipeGPT) for recipe generation using Recipe1M dataset.\\

Majumder et al. generated the personalized recipes based on user preference using dataset of 180k recipes and 700k user interactions. Authors used bidirectional encoder-decoder to generate recipes and evaluate the generated recipes. In a similar spirit, a routing enforced generative model~\cite{yu2020routing} was proposed to generate recipes considering the ingredients and user preference. Transformer-based models such as GPT\cite{radford2019language}, BERT\cite{devlin2018bert}, Roberta\cite{liu2019roberta} outperform the LSTM \cite{greff2016lstm} and RNN-based models to generate the novel recipes. Recent study~ \cite{zhang2019dialogpt} showed that to generate recipes, OpenAI's GPT2 model outperforms the state-of-the-art models. 
\\

The limitations that we draw from the previous literature is that the recipes generated by RecipeGPT and RecipeNLG are not well structured and taking more processing time in generating a recipe. Our aim is to resolve these issue and upgrade our model so that it will generate a new recipe within lesser time and used special tokens to account the fractions and numbers. 

\section{Dataset}
\begin{figure}[!htb]
\includegraphics[width=0.5\textwidth]{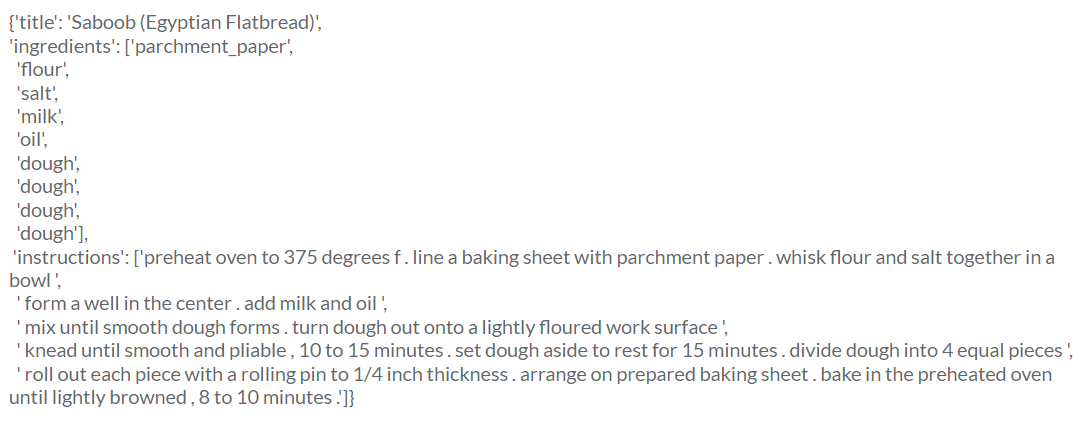}
\caption{Dataset before preprocessing}
\label{before_data}
\end{figure}

The model is trained on the RecipeDB dataset~\cite{batra2020recipedb} (see Fig.~\ref{before_data}) to generate novel recipes. RecipeDB is a structured compilation of recipes, ingredients and nutrition profiles interlinked with flavor profiles and health associations. The repertoire comprises of 118,171 recipes from cuisines across the globe (6 continents, 26 geo-cultural regions and 74 countries), cooked using 268 processes (heat, cook, boil, simmer, bake, etc.), by blending over 20,262 diverse ingredients which are further linked to their flavor molecules (FlavorDB~\cite{garg2018flavordb}), nutritional profiles (USDA) and empirical records of disease associations obtained from Medline (DietRx). This resource is aimed at facilitating scientific explorations of the culinary space (recipe, ingredient, cooking processes, dietary styles) to taste attributes (flavor profile) and health (nutrition and disease associations) seeking for divergent applications.\\

\begin{figure}[!htb]
\includegraphics[width=0.5\textwidth]{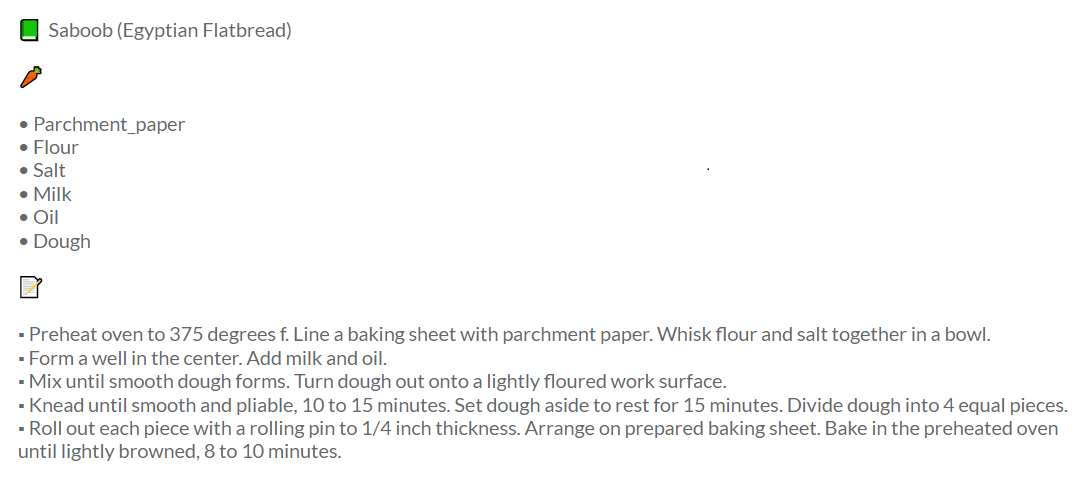}
\caption{Dataset after preprocessing}
\label{after_data}
\end{figure}

To improve the quality of generated recipes, the dataset is further preprocessed which involves removing incomplete and redundant recipes, fixing the length of recipes to 2000 characters as on plotting recipe size distribution it is seen that most of the recipes covers the range of 2000 characters. We convert the cooking recipes in a specific format containing title, ingredients and instructions (see Fig.~\ref{after_data}). 

\section{Recipe Generation Models}
The goal was to have a model which should be capable of generating a novel cooking recipe given ingredients (or type of food processing) as input. For solving the above problem statement we experimented with various text generation models. The models were trained/fine-tuned on RecipeDB dataset. The generated recipes were evaluated at odds to the original recipe using the evaluation metrics.

\subsection{LSTMs}
We experimented with a LSTM model as our baseline. The LSTM model was trained with both character level as well as word level on the RecipeDB dataset. For each character or word, the model looks up the embedding and applies the dense layer to generate logits which predicts the log-likelihood of next character or word. The problem with the generated recipes was that LSTM based models were able to learn only the structure of the recipes, not the in-depth relationship between the ingredients and recipe preparatory instructions as LSTM based models are often prone to over-fitting and it was a big overhead to apply a proper dropout algorithm to curb this problem.   

\subsection{GPT-2}
Transformers are one of the leading architectures in language modeling tasks. We have used Generative Pretrained Transformer 2 (GPT2)~\cite{radford2019language}, an open-source transformer model from OpenAI, for conditional text generation. The model uses attention mechanism, which is currently the principal component in any state-of-the-art transformer model. We experimented with few other implementations and variations of GPT2. The model provided by Hugging Face (\url{https://huggingface.co/}) was used as the base model for training our language model. 

\begin{figure}[!htb]
\includegraphics[width=0.5\textwidth]{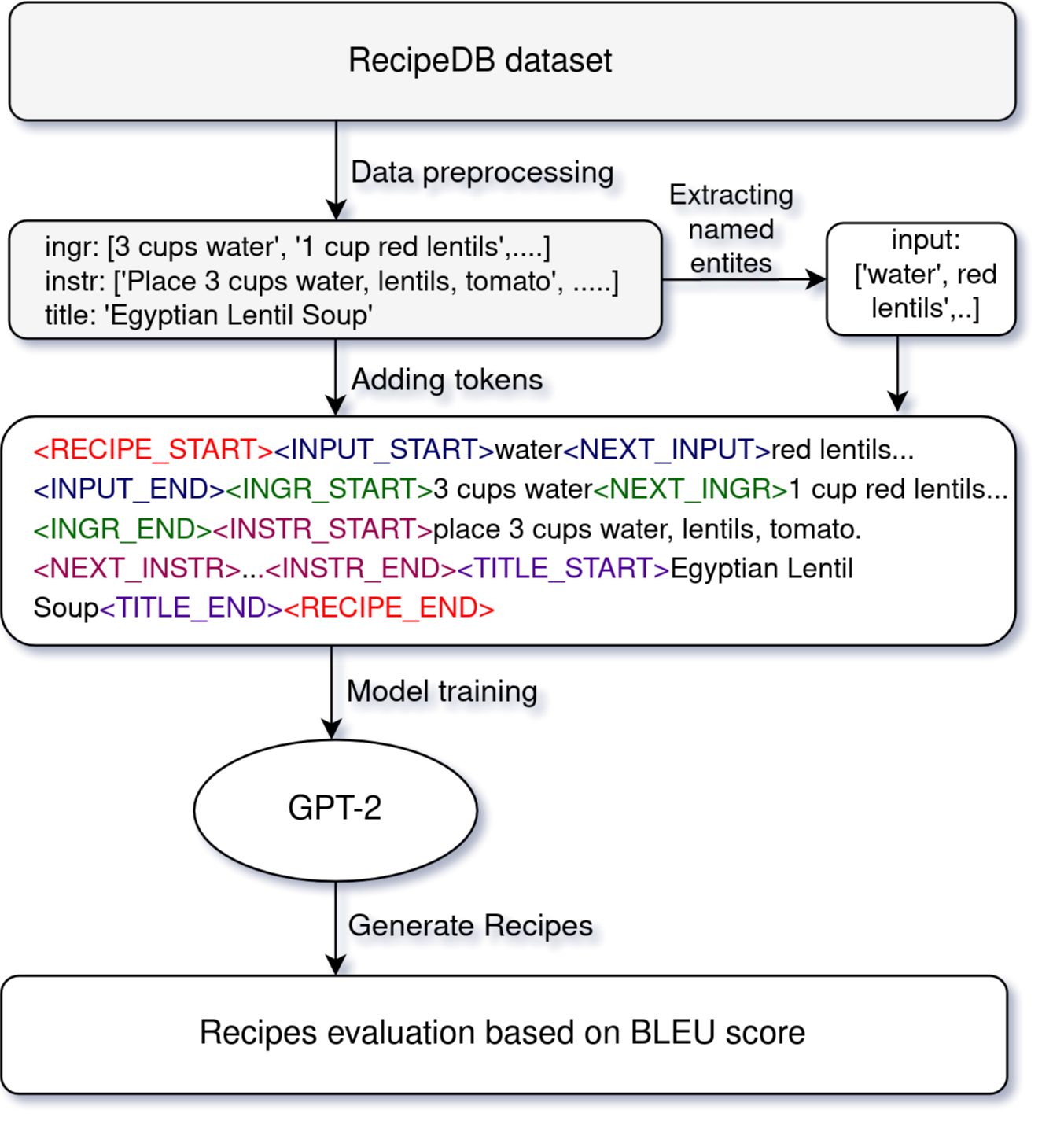}
\caption{Flow diagram of recipe generation}
\label{schema}
\end{figure}

To generate recipes our language model takes ingredients as an input. For the model to be able to understand the actual structure of a food recipe we preprocess the data in such a way that it is one long string with all the recipes with different tags that differentiate between different sections of the recipe as shown in Fig.~\ref{schema} while fine-tuning the GPT2 model. While training, recipe elements present in RecipeDB along with a vanilla (random) recipe was used as a single training instance, to speedup the training process. We have considered approximately 2$\sigma$ (95.46 percent) in recipe size distribution curve of RecipeDB dataset. Few short length recipes ($-3\sigma$) were merged to make the length close to the mean length of the recipe size distribution curve.

\section{Results}
For evaluating the recipes, we calculated the BLEU score---higher BLEU score indicates generated text is closer to the actual reference text. Apart from other models, GPT2 model maintains the contextual information of the sentence and hence give good results. We used an Nvidia A100 Tensorcode GPU server from our own lab which speed up the training process. On CPU, it’s taking 2-3 days to train our whole model but on GPU it took around 16 hours to train the model. We have used different dataset and preprocessing techniques, so we can't compare our results with state-of-the-art models in terms of BLEU score. Table~\ref{tab:perfstat} shows the performance statistics of applied LSTM and GPT2 model on RecipeDB dataset.

\begin{table}[!htb]
    \centering
    \caption{Performance Statistics of Models}
    \label{tab:perfstat}
    \begin{tabular}{|l|c|}
    \hline
    \textbf{Model} & \textbf{BLEU Score} \\
    \hline
    Char-level LSTM & 0.347\\
    \hline
    Word-level LSTM & 0.412\\
    \hline
    DistilGPT2 & 0.442\\
    \hline
    \textbf{GPT-2 medium} & \textbf{0.806}\\
    \hline
    \end{tabular}
   
\end{table}

Deep learning models do not require any human intervention or task-specific algorithms to extract the features. GPT-2 is trained on eight million text documents. The dataset is scraped from the web and it contains the recipes from different regions, the dataset is unorganised and needed more manual preprocessing.

\section{Web Application}
\begin{figure}[!htb]
\includegraphics[width=0.5\textwidth]{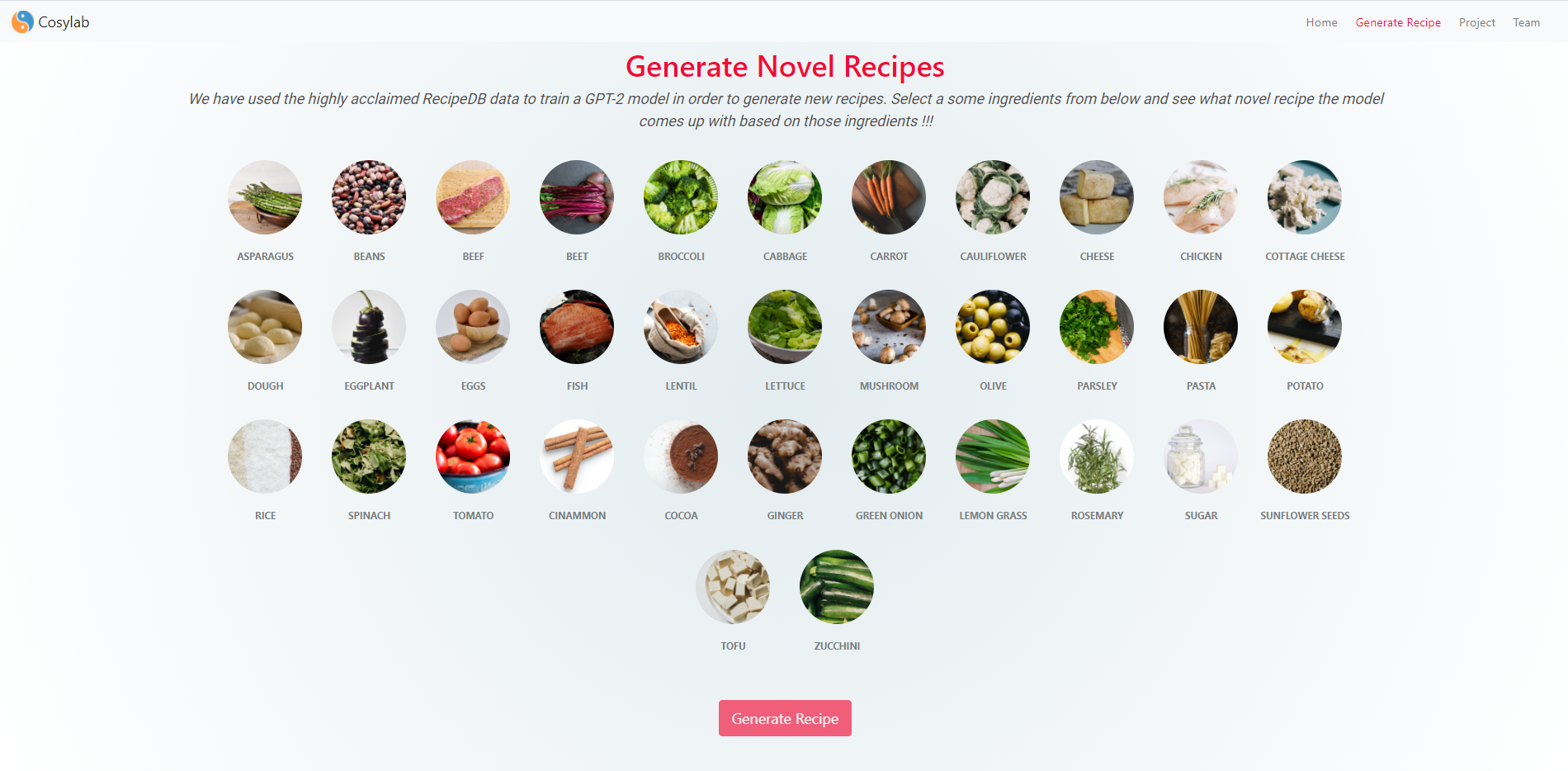}
\caption{Website interface to choose ingredients and generate recipe}
\label{website}
\end{figure}

\begin{figure}[!htb]
\includegraphics[width=0.5\textwidth]{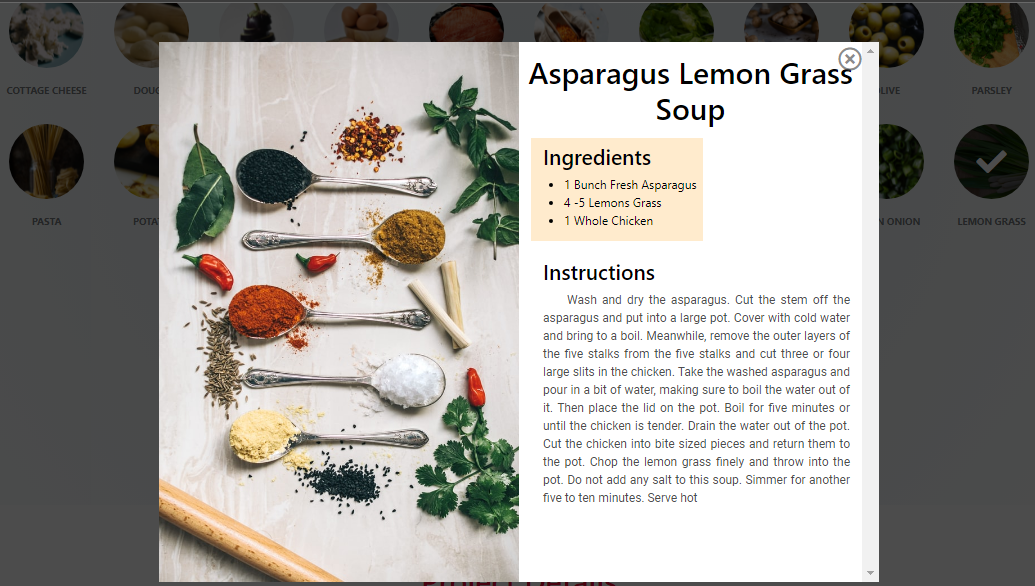}
\caption{Recipe Generated using GPT2 model}
\label{gen_recipe}
\end{figure}

Web applications are one of the best ways to demonstrate the work. The goal of this web application is to create a frontend where user provides the ingredients (see Fig.~\ref{website}). In the backend, proposed generative model processes the input and generates the recipes with title, ingredients and instructions (see Fig.~\ref{gen_recipe}).

The architecture of web application is built on Python framework. For the frontend, ReactJS is used which is very simple and lightweight library and Flask is used for backend. To handle more user requests and prevents breakage of application, frontend is completely decoupled from the backend using microservices architecture. The frontend and backend are separately dockerized and hosted on the webserver. In future, if load increase then developer only need to replicate the docker.

\section{Conclusion and Future Work}
In this work, we used neural-network based LSTM model and transformer-based GPT2 model to generate the novel recipes using list of ingredients as input. We accounted the ingredients quantity in each recipe which was not present in earlier studies~\cite{bien2020recipenlg}. Transformer-based GPT2 model outperforms the neural-network based LSTM model with BLEU score of 0.806. The major shortcoming that we faced is evaluating the generated recipes. For future work, we intend to use GPT-Neo which is built on similar architecture of GPT-3. The main challenge faced during this research was resources and hardware limitations. We have limited hours of GPU, RAM and Disk space on Google Colab, which lead to session crashing after every 5 to 7 epochs. 

\section{Acknowledgement}
G.B. thanks Indraprastha Institute of Information Technology (IIIT Delhi) for the computational support. G.B. thanks Technology Innovation Hub (TiH) Anubhuti for the research grant. V.P., P.C., M.K., S.T., A.S. are M.Tech students and M.G. is a research scholar in Dr. Bagler's lab at IIIT Delhi and thankful to IIIT Delhi for the support. M.G. thanks IIIT Delhi for the fellowship. We would like to thank Ankur Goel for his support in website deployment. 

\bibliographystyle{IEEEtran}
\bibliography{ratatouille}

\end{document}